\pdfoutput=1

\documentclass[11pt]{article}

\usepackage[]{acl}

\usepackage{times}
\usepackage{latexsym}

\usepackage[T1]{fontenc}

\usepackage[utf8]{inputenc}

\usepackage{microtype}
\usepackage{xcolor}
\usepackage{color}
\usepackage{tcolorbox}
\usepackage{CJK}
\usepackage{adjustbox}
\usepackage{float}

\definecolor{melon}{RGB}{248,158,123}
\definecolor{salmon}{RGB}{246,146,137}
\definecolor{yellowgreen}{RGB}{152,204,112}

\usepackage{multirow}
\usepackage[normalem]{ulem}
\useunder{\uline}{\ul}{}
\usepackage{tabularx}

\usepackage{graphicx}

\usepackage{amssymb}
\usepackage{amsmath}
\usepackage{bm}
\usepackage{bbm}
\usepackage{algorithm2e}[ruled]
\usepackage{algpseudocode}
\usepackage{enumitem}

\title{Examining the Causal Effect of First Names on Language Models:\\ The Case of Social Commonsense Reasoning}

\author{Sullam Jeoung \ \ \  Jana Diesner \ \ \  Halil Kilicoglu\\
University of Illinois at Urbana-Champaign\\
  \texttt{\{sjeoung2, jdiesner ,halil\}@illinois.edu}}

\begin{document}
\maketitle
\begin{abstract}
As language models continue to be integrated into applications of personal and societal relevance, ensuring these models' trustworthiness is crucial, particularly with respect to producing consistent outputs regardless of sensitive attributes. Given that first names may serve as proxies for (intersectional) socio-demographic representations, it is imperative to examine the impact of first names on commonsense reasoning capabilities. In this paper, we study whether a model's reasoning given a specific input differs based on the first names provided. Our underlying assumption is that the reasoning about \textit{Alice} should not differ from the reasoning about \textit{James}. We propose and implement a controlled experimental framework to measure the causal effect of first names on commonsense reasoning, enabling us to distinguish between model predictions due to chance and caused by actual factors of interest. Our results indicate that the frequency of first names has a direct effect on model prediction, with less frequent names yielding divergent predictions compared to more frequent names. To gain insights into the internal mechanisms of models that are contributing to these behaviors, we also conduct an in-depth explainable analysis. Overall, our findings suggest that to ensure model robustness, it is essential to augment datasets with more diverse first names during the configuration stage.  
\end{abstract}

\section{Introduction}
\begin{figure}[]
\centerline{\includegraphics[width=\columnwidth, height=2.3in]{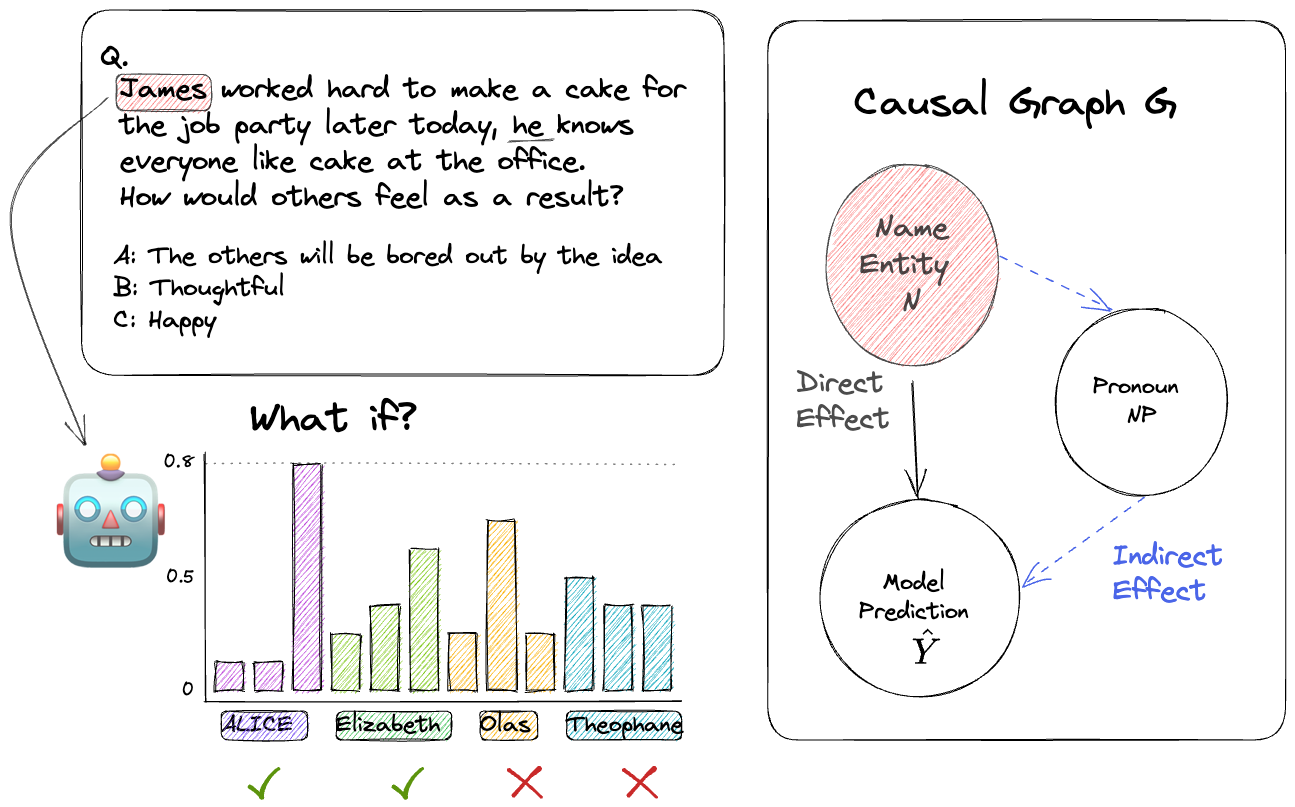}}
\caption{Framework of our approach. (Left): An example template with name instances (Right): The causal graph $G$ we hypothesize for analysis}
\label{fig:drawing}
\end{figure}

Recent language models (LMs) \cite{brown2020language,radford2019language} have shown remarkable improvements when used in NLP tasks and are increasingly used across various application domains to engage with users and address their personal and social needs, such as AI-assisted autocomplete and counseling \cite{hovy2021importance,sharma2021towards}. As these LMs models are adopted, their social intelligence and commonsense reasoning have become more important, especially as AI models are deployed in situations requiring social skills \cite{wang2007social,wang2019persuasion}.\\
In this paper, we examine how first names are handled in commonsense reasoning (Fig \ref{fig:drawing}). To this end, we measure the causal effect that name instances have on LMs' commonsense reasoning abilities. A key aspect of commonsense reasoning of LMs should be that they provide consistent responses regardless of the subject's name or identity \cite{sap2019social}. That is, the reasoning behind "\textit{Alice}" should not differ from that about "\textit{James}", for instance. Given that first names can be a proxy for representation of gender and/ or race, this consistency is essential not only for the robustness but also for the fairness and utility of a LM.\\
Previous studies have revealed that pre-trained language models are susceptible to biases related to peoples' first names. For instance, in the context of sentiment analysis, certain names have been consistently associated with negative sentiments by language models \cite{prabhakaran2019perturbation}. Additionally, during text generation, names have been found to be linked to well-known public figures, indicating biased representations of names \cite{shwartz2020you}. Furthermore, \citet{wolfe2021low} demonstrated that less common names are more likely to be ‘subtokenized’ and associated with negative sentiments compared to frequent names. These studies shed light on how pre-trained language models disproportionately process name representations, potentially leading to biased outputs.\\
While examining pre-trained language models is valuable to understand their capabilities and limitations, in many cases the models are fine-tuned, or adapted and optimized, to guarantee improved performance on specific downstream tasks, such as text classification, machine translation, and question answering, among others \cite{bai2004using,peng2007large,rajpurkar2018know}.\\
Given that fine-tuning pre-trained language models can lead to major performance gains \cite{devlin2019bert}, in this paper, we ask if performance disparities based on names still exist even when the models are fine-tuned. If so, we ask which components of the models contribute to performance disparities and to what extent. We design a controlled experimental setting to determine whether performance differences arise by chance or are caused by names. Our contributions are three-fold\footnote{The source code is available: \url{https://github.com/sullamij/Causal-First-Names/}}: 
\begin{itemize}[noitemsep,topsep=0pt]
    \item We propose a controlled experimental framework based on a causal graph to discern the causal effect of first names in the commonsense reasoning of language models. We leverage the name statistics from U.S. Census data for this purpose. 
    \item We present an in-depth analysis to understand the internal model mechanisms in processing first names. To be specific, we examine the embeddings and neuron activation of first names. 
    \item Based on our analysis, we provide suggestions for researchers in configuring the datasets to provide more robust language modeling. 
\end{itemize}
\section{Task Formulation}  
\label{sec:formalization}
We consider a dataset of commonsense reasoning examples $d \in \mathcal{D}$, where each item consists of a question $q \in \mathcal{Q}$, three possible answer candidates $\mathcal{C}=\{c_1,c_2,c_3\}$, and a label $y \in Y$, which is the correct answer among the candidates. $\mathcal{Q}$ and $\mathcal{C}$ serve as a template $\bm{t}$, containing placeholders for names $[\bm{n}]$ and pronouns referring to the names, $[\bm{np}]$.\\ 
To ensure grammatical correctness, a pronoun placeholder $\bm{np}$ is set in variants of subject pronoun $\bm{np_{1}}$, object pronoun $\bm{np_{2}}$, and dependent possessive pronouns $\bm{np_{3}}$. An example of the data template is as follows: \\\newline 
\setlength{\fboxrule}{0.4pt}
\noindent\fbox{%
    \parbox{\columnwidth}{%
\textbf{Question $\mathcal{Q}$}: Typically every four months, \colorbox{melon}{$[n]$} went to the doctor for a routine checkup and was told \colorbox{yellowgreen}{$[{np_1}]$} needs rest. What will \colorbox{melon}{$[{n}]$} want to do next?\\  \linebreak
\textbf{Candidates $\mathcal{C}$}:\{ \\  
\textbf{(a)} call the doctor, \textbf{(b)} finish all \colorbox{yellowgreen}{$[{np_{2}}]$} projects and postpone the rest, 
 \textbf{(c)} take time off from work\}\ \linebreak
\textbf{Label $y$}: \textbf{(c)} take time off from work\\
}%
}

\section{Causal Graph}
A language model can be denoted as a function $f$, taking inputs as follows:
\begin{align}
    \hat{y}=f(t(\bm{n},\bm{np}))
\end{align}
We are interested in how first names ($\bm{n} \in N$) influence the prediction $\hat{y} \in \hat{Y}$ under the function $f$. 
We hypothesize that there is a causal graph $\mathcal{G}$ that encodes possible causal paths relating first names to the model's prediction (Fig \ref{fig:drawing}, right). \footnote{Specifically, when referring to the causal graph, it pertains to the utilization of causal directed-acyclic graphs (DAGs), as mentioned in the work by \cite{feder2022causal}}\\
We identify both the direct effect and indirect effect on model prediction \cite{pearl2022direct}:

1. The direct effect of names on model prediction $(N \rightarrow \hat{Y})$ measures how names have a direct impact on model predictions (without going through any intermediate variables).

2. The indirect effect indicates potential confounding factors associated with names that may influence predictions. We hypothesize that pronouns are an intermediate variable $(N \rightarrow NP \rightarrow \hat{Y})$. Intuitively, pronouns that refer to names can influence how models make their predictions. For example, this indirect effect indicates changes in model prediction when the pronouns differ (e.g. \textit{he} vs. \textit{she}) but the names remain the same or fixed (e.g. \textit{Pat}). Pronouns inherently associate with the names they refer to, and this association may cue models to consider those names more strongly when generating a response. Thus, we posit the effect of pronouns as an indirect effect.\\\newline   
Below, we formalize the causal mechanisms, intervention lists, and the effect size that measures the change in model prediction.\\\newline
\textbf{Direct Effect} 
\begin{align*}
\text{DE} (N \rightarrow \hat{Y}) &:= \\ \sum_t &\mathbb{E}_N^+[\hat{Y}|T=t] - \mathbb{E}_N^-[\hat{Y}|T=t]    
\end{align*}
\newline
where $\mathbb{E}_{N}^+[\hat{Y}|T=t]$ indicates the average effect size of name lists $N^+$, while $\mathbb{E}_{N}^-[\hat{Y}|T=t]$ indicates the average effect size of name lists $N^-$ on template $t$. The details of the name lists of interest $N^+$ and $N^-$ are listed in section \ref{causalintervention} and the effect size is defined in section \ref{sec:effectsize}. $\text{DE}$ measures the causal effects between name lists via direct \texttt{do-}interventions of $N^+$ as the template $t$ is fixed \cite{pearl1995causal}. Beyond computing the differences, to test the null hypothesis, we conduct a \textit{t}-test and obtain the \textit{p}-value statistics. \\\newline
\textbf{Indirect effect}
\begin{align*}
    \text{IE} (N \rightarrow \hat{Y}) :=   \sum_t^T\sum_n^N & (\mathbb{E}_{NP}^+[\hat{Y}|T=t,N=n] \\ & - \mathbb{E}_{NP}^-[\hat{Y}|T=t,N=n])
\end{align*}
\newline
where $\mathbb{E}^+_{NP}[\hat{Y}|T=t,N=n]$ indicates the average prediction conditioned on template $t$ and name $n$, with the set of $NP^+$, and $\mathbb{E}^-_{NP}[\hat{Y}|T=t,N=n]$ refers that of $NP^-$. To account for the effect of names, note that names are also controlled along with the template.  
\subsection{Causal Intervention}
\label{causalintervention}
We apply feasible intervention on  $T:\{q,c,(n,np),y\}$ to $T':\{q,c,(n',np'),y\}$. We denote the intervention list as \texttt{Do}$(X:x \rightarrow x')$, where $X \in \{\mathcal{Q},\mathcal{C},(N,NP),Y\}$. We denote $\hat{y}' \in \hat{Y}'$ to indicate the prediction of the intervened $X'$.
As we want to explore names based on their characteristics, we partition the intervention lists $N$ based on two criteria: \textit{frequency} and \textit{gender}. These criteria were chosen following previous work \cite{wolfe2021low,buolamwini2018gender} that has demonstrated that less common names, as well as gender, can be key factors in models that exhibit biases. Studies have shown that models trained on datasets with an imbalance of names or gender can reflect and even amplify prejudices, resulting in unfair outcomes, particularly for marginalized groups \cite{bolukbasi2016man,zhao2017men}. By focusing on name frequency and gender representation, we aim to evaluate the impact of these criteria on models.\\
In order to base our work on prior statistics, we use the name statistics from the U.S. Census data. The detailed process of how the intervention list was filtered from the dataset is outlined in section \ref{sec:experiment details}. We consider the set of names for $\texttt{do-intervention}$ as below:\\\newline
\textbf{\textsc{Most-Least}} Based on the frequency of names, $N_\textsc{Most}$ indicates the names with top-${k}$ highest frequency, whereas $N_\textsc{Least}$ refers to lowest frequency. \\\newline
\textbf{\textsc{Female-Male}} We use the gender information from the statistics to discern the gender of a name. Note that we purely refer to the `gender' of names based on their records. That is, we account for cases where a name can be both male or female, based on the frequency statistics. For example, if the records for \textit{Lee} exist for both males and females, we consider the name belonging to both genders to reflect real-world data.\\

\subsection{Effect Size} 
\label{sec:effectsize}
To evaluate the impact of our model, we utilize two distinct metrics.\newline
\textbf{\textsc{Accuracy}} To quantify the degree of wrong predictions, we define $\mathbf{d}_{\textsc{Acc}}$ as
$$\mathbf{d}_{\textsc{ACC}}(x):=\mathbbm{1}(\hat{y}\neq y)$$
\begin{align*}
    \mathbf{d}_{\textsc{ACC}}(X' \rightarrow X) = \frac{\mathbf{d}_{\textsc{ACC}}(X')-\mathbf{d}_{\textsc{ACC}}(X)}{\mathbf{d}_{\textsc{ACC}}(X)}
\end{align*}
\textbf{\textsc{Agreement}} This metric measures the extent to which the model's predictions vary in response to different interventions. The rationale behind this metric stems from the recognition that the task under consideration entails a multiple-choice problem. Additionally, in real-world scenarios, it is often the case that a definitive 'ground truth' may not exist. Consequently, we employ this metric to measure the divergence of predictions. This metric goes beyond simple accuracy, which merely determines the correctness or incorrectness of predictions. Instead, this objective is to evaluate the diversity of predictions, thereby taking into consideration the range of errors that may arise. To calculate the \textbf{\textsc{Agr}} score, which is a modification of Fleiss' kappa \cite{fleiss1973equivalence}, we begin with a list of $N$ names and obtain a score: 
\begin{align*}
    \mathbf{d}_{\textsc{AGR}}(X)=\frac{1}{|N|\cdot|N-1|}\sum_{j=1}^k(n_j\cdot(n_j-1))
\end{align*}
\begin{align*}
    \mathbf{d}_{\textsc{AGR}}(X' \rightarrow X) = \frac{\mathbf{d}_{\textsc{AGR}}(X')-\mathbf{d}_{\textsc{AGR}}(X)}{\mathbf{d}_{\textsc{AGR}}(X)}
\end{align*}
where $|N|$ indicates the total number of names in name lists, $k$ the number of categories (e.g. in our case, $k=3$, \{(a),(b),(c)\}), and $n_j$ the number of instances predicting the answer as category $j$. The \textbf{\textsc{Agr}} score ranges from 0 to 1, with a score of 1 indicating complete agreement among all name instances in their category prediction, and a score of 0 indicating no agreement. This metric enables us to assess the degree to which a model's predictions are sensitive to different interventions. 

\section{Explanations of Causal Effects}\label{sec:Explanation}
The causal analysis shows the surface-level comparison of model outputs but fails to capture the nuanced processes underlying each model's reasoning.  By probing the internal workings of the models, we seek to gain insights into how the models derive their conclusions and also where their approaches diverge. We use two approaches to gain a deeper understanding of the models' predictions. First, we analyze the models' internal representations to discern how they encode various names. Specifically, we focus on the distinction in contextualization between the embeddings of frequent names and less frequent names. Second, we apply a diagnostic technique based on neuron activation to pinpoint how the models process names. 

\subsection{Contextualization of Name Representations}
We investigate the contextualization of name representations in language models with respect to their characteristics. We partition the names based on frequency \textsc{Most} and \textsc{Least} and compare the degree of contextualization. To be specific, we measure the similarity between name representations at each layer of the model by following the approach proposed by \citet{wolfe2021low}. In order to ensure that the embeddings being compared are based on the same space, we restrict the comparison to representations within each layer and do not compare across different layers. We adopt two commonly used metrics to validate the overall trend observed in our analysis.\\\newline
\textbf{\textsc{Cosine Similarity}} \ The cosine-similarity of name $w$, in layer $l$ is formalized as followes: $$c(\mathbf{w})_l = \frac{1}{n^2-n}\sum_i\sum_{j \neq i}cos(\vec{w}_i,\vec{w}_j)$$ where $n$ refers to the total number of name pairs. This corresponds to the self-similarity studied in \cite{ethayarajh2019contextual,wolfe2021low}. The measure lies ranges from 0 to 1, where 1 indicates high similarity, and 0 otherwise.\\
\textbf{\textsc{Linear CKA}} (Centered Kernel Alignment) This similarity metric measures similarity in neural network representations and was proposed by \citet{kornblith2019similarity}. It ranges from 0 to 1, where 1 indicates perfect similarity, and 0 otherwise. $$\frac{||\mathbf{x_j}^{\top}\mathbf{x_i}||_F^2}{||\mathbf{x_i}^\top \mathbf{x_i}||_F ||\mathbf{x_j}^\top \mathbf{x_j}||_F}$$ where $\mathbf{x_i}$ and $\mathbf{x_j}$ indicates two randomly selected name embeddings, such that $i\neq j$.

\begin{table*}[!ht]
{
\centering
\begin{tabular}{lccc|ccc}
\hline
\multicolumn{1}{c|}{\multirow{2}{*}{\begin{tabular}[c]{@{}c@{}}Effect size: $\mathbf{d}_{ACC}$\\ (Accuracy)\end{tabular}}} 
  & \multicolumn{3}{c|}{\begin{tabular}[c]{@{}c@{}}\textbf{Not-finetuned}\\ \textit{(Epoch 0)}\end{tabular}} & \multicolumn{3}{c}{\begin{tabular}[c]{@{}c@{}}\textbf{Fine-tuned} \\ \textit{(Epoch10)}\end{tabular}}\\ \cline{2-7} \multicolumn{1}{l|}{} 
 & \textsc{Gpt2}  & \textsc{Bert}  & \textsc{RoBERTa}                              & \textsc{Gpt2}  & \textsc{Bert} & \textsc{RoBERTa}                                                           \\ \hline
\multicolumn{1}{c|}{\textsc{Most} $\rightarrow$ \textsc{Least}} & \begin{tabular}[c]{@{}c@{}}-.07\\ $^{(.354)}$\end{tabular}   & \begin{tabular}[c]{@{}c@{}}$\mathbf{.258^{***}}$\\ $^{(<.001)}$\end{tabular} & \begin{tabular}[c]{@{}c@{}}-.04\\ $^{(.534)}$\end{tabular}                & \begin{tabular}[c]{@{}c@{}}.002\\ $^{(.956)}$\end{tabular}           & \begin{tabular}[c]{@{}c@{}}.007\\ $^{(.841)}$\end{tabular}              & \begin{tabular}[c]{@{}c@{}}.004\\ $^{(.884)}$\end{tabular} \\
\multicolumn{1}{c|}{\textsc{Male} $\rightarrow$ \textsc{Female}}              & \begin{tabular}[c]{@{}c@{}}.001\\ $^{(.801)}$\end{tabular}   & \begin{tabular}[c]{@{}c@{}}.005\\ $^{(.634)}$\end{tabular}               & \begin{tabular}[c]{@{}c@{}}-.025\\ $^{(.627)}$\end{tabular}               & \begin{tabular}[c]{@{}c@{}}.002\\ $^{(.819)}$\end{tabular}           & \begin{tabular}[c]{@{}c@{}}-.002\\ $^{(.965)}$\end{tabular}             & \begin{tabular}[c]{@{}c@{}}.002\\ $^{(.751)}$\end{tabular}           \\
\multicolumn{1}{c|}{\begin{tabular}[c]{@{}l@{}}\textsc{Most Male} $\rightarrow$ \textsc{Least Male}\end{tabular}}     & \begin{tabular}[c]{@{}c@{}}-.059\\ $^{(.365)}$\end{tabular}  & \begin{tabular}[c]{@{}c@{}}$\mathbf{.275}^{***}$\\ $^{(<.001)}$\end{tabular}  & \begin{tabular}[c]{@{}c@{}}-.018\\ $^{(.627)}$\end{tabular}               & \begin{tabular}[c]{@{}c@{}}-.004\\ $^{(.906)}$\end{tabular}          & \begin{tabular}[c]{@{}c@{}}.006\\ $^{(.885)}$\end{tabular}              & \begin{tabular}[c]{@{}c@{}}.011\\ $^{(.751)}$\end{tabular} \\
\multicolumn{1}{c|}{\begin{tabular}[c]{@{}l@{}}\textsc{Most Female} $\rightarrow$ \textsc{Least Female}\end{tabular}} & \begin{tabular}[c]{@{}c@{}}-.089\\ $^{(.349)}$\end{tabular}  & \begin{tabular}[c]{@{}c@{}}$\mathbf{.241}^{***}$\\ $^{(<.001)}$\end{tabular}  & \begin{tabular}[c]{@{}c@{}}-.06\\ $^{(.800)}$\end{tabular}                & \begin{tabular}[c]{@{}c@{}}.008\\ $^{(.990)}$\end{tabular}           & \begin{tabular}[c]{@{}c@{}}.008\\ $^{(.800)}$\end{tabular}              & \begin{tabular}[c]{@{}c@{}}-.002\\ $^{(.954)}$\end{tabular}    \\\hline
\end{tabular}}
\caption{Direct Effect: Accuracy ($\mathbf{d_{\textsc{Acc}}}$) score of the models with and without fine-tuning. The numbers in parentheses are \textit{p-values}. The values in bold indicate the significant effects with \textit{p-values}$<0.05$. The results show that after fine-tuning, the direct effects are not significant.}
\label{tab:direct-acc}
\end{table*}

\begin{table*}[!ht]
{
\centering
\begin{tabular}{lccc|ccc}
\hline
\multicolumn{1}{c|}{\multirow{2}{*}{\begin{tabular}[c]{@{}c@{}}Effect size: $\mathbf{d}_{AGR}$\\ (Agreement)\end{tabular}}} 
  & \multicolumn{3}{c|}{\begin{tabular}[c]{@{}c@{}}\textbf{Not-finetuned}\\ \textit{(Epoch 0)}\end{tabular}} & \multicolumn{3}{c}{\begin{tabular}[c]{@{}c@{}}\textbf{Fine-tuned} \\ \textit{(Epoch10)}\end{tabular}}\\ \cline{2-7} \multicolumn{1}{l|}{} 
 & \textsc{Gpt2}  & \textsc{Bert}  & \textsc{RoBERTa}                              & \textsc{Gpt2}  & \textsc{Bert} & \textsc{RoBERTa}                                                           \\ \hline
 \multicolumn{1}{c|}{\textsc{Most} $\rightarrow$ \textsc{Least}}               & \begin{tabular}[c]{@{}c@{}}-.0004\\ $^{(.954)}$\end{tabular} & \begin{tabular}[c]{@{}c@{}}$\mathbf{.058}^{***}$\\ $^{(<.001)}$\end{tabular}     & \begin{tabular}[c]{@{}c@{}}$\mathbf{.048}^{***}$\\ $^{(<.001)}$\end{tabular}     & \begin{tabular}[c]{@{}c@{}}$\mathbf{.013}^{*}$\\ $^{(.02)}$\end{tabular}  & \begin{tabular}[c]{@{}c@{}}$\mathbf{.022}^{***}$\\ $^{(<.001)}$\end{tabular}   & \begin{tabular}[c]{@{}c@{}}$\mathbf{.012}^*$\\ $^{(.02)}$\end{tabular} \\
\multicolumn{1}{c|}{\textsc{Male} $\rightarrow$ \textsc{Female}}              & \begin{tabular}[c]{@{}c@{}}.02\\ $^{(.712)}$\end{tabular}    & \begin{tabular}[c]{@{}c@{}}.009\\ $^{(.306)}$\end{tabular}               & \begin{tabular}[c]{@{}c@{}}.010\\ $^{(.267)}$\end{tabular}                & \begin{tabular}[c]{@{}c@{}}.004\\ $^{(.565)}$\end{tabular}           & \begin{tabular}[c]{@{}c@{}}-.002\\ $^{(.722)}$\end{tabular}             & \begin{tabular}[c]{@{}c@{}}.007\\ $^{(.354)}$\end{tabular}  \\ 
\multicolumn{1}{c|}{\begin{tabular}[c]{@{}l@{}}\textsc{Most Male} $\rightarrow$ \textsc{Least Male}\end{tabular}}     & \begin{tabular}[c]{@{}c@{}}.003\\ $^{(.748)}$\end{tabular}   & \begin{tabular}[c]{@{}c@{}}$\mathbf{.068}^{***}$\\ $^{(.0)}$\end{tabular}     & \begin{tabular}[c]{@{}c@{}}$\mathbf{.060}^{***}$\\ $^{(<.001)}$\end{tabular} & \begin{tabular}[c]{@{}c@{}}$\mathbf{.017}^*$\\ $^{(.028)}$\end{tabular} & \begin{tabular}[c]{@{}c@{}}$\mathbf{.027}^{***}$\\ $^{(<.001)}$\end{tabular} & \begin{tabular}[c]{@{}c@{}}.015\\ $^{(.052)}$\end{tabular} \\
\multicolumn{1}{c|}{\begin{tabular}[c]{@{}l@{}}\textsc{Most Female} $\rightarrow$ \textsc{Least Female}\end{tabular}}& \begin{tabular}[c]{@{}c@{}}-.004\\ $^{(.691)}$\end{tabular}  & \begin{tabular}[c]{@{}c@{}}$\mathbf{.047}^{**}$\\ $^{(.004)}$\end{tabular}    & \begin{tabular}[c]{@{}c@{}}$\mathbf{.03}^{***}$\\ $^{(<.001)}$\end{tabular}    & \begin{tabular}[c]{@{}c@{}}.009\\ $^{(.262)}$\end{tabular}           & \begin{tabular}[c]{@{}c@{}}.016\\ $^{(.240)}$\end{tabular}              & \begin{tabular}[c]{@{}c@{}}$\mathbf{.010}^*$\\ $^{(.036)}$\end{tabular}\\ \hline
\end{tabular}
}
\caption{Direct Effect: Agreement ($\mathbf{d_{\textsc{Agr}}}$) score of the models with and without fine-tuning. The numbers in parentheses are \textit{p-values}. The values in bold indicate the significant effects with \textit{p-values}$<0.05$. The results show that after being fine-tuned, the effects show significance in the frequency of the names (row1). The asterisks indicate the significance level: ($^{***} p \leq 0.001$,$^{**} p \leq 0.01$, $^{*} p \leq 0.05$) }
\label{tab:direct-agr}
\end{table*}

\subsection{Neuron Activations}
Previous work has explored the activation patterns of neurons in deep neural networks for the domains of language and vision as a means of gaining insight into the inner workings of such networks \cite{karpathy2015visualizing,
 poerner2018interpretable, olah2018building, dalvi2019neurox}. It has been demonstrated that the feed-forward network (FF) component of transformer architectures encodes a significant amount of information \cite{wang2022finding,geva2021transformer}. Building on this prior work, we conducted a detailed analysis of how neuron activations vary according to different characteristics of the input data. Our analysis involved extracting the activations of the FF network's neurons based on the hidden states of previous layers and applying non-negative matrix factorization (NMF) \cite{cichocki2009fast} to decompose these activations into semantically meaningful components. By visualizing groups of neuron activations, we aim to gain a better understanding of the models' internal mechanisms, and how the models construct their representations and predictions. For the detailed algorithm see Appendix \ref{sec:algorithm} outlines the steps involved in this analysis. 

\section{Experimental Setup}
\label{sec:experiment details}
\textbf{Dataset} We use the \textsc{SocialIQA} dataset from \citet{sap2019social}. The selection of this dataset is motivated by its suitability for investigating model behavior in a social context, as the dataset consists of questions for probing \textit{emotional} and \textit{social} intelligence in everyday situations. 
By analyzing the model's responses to questions pertaining to social and emotional intelligence, valuable insights can be gleaned regarding the models' handling of some nuances of human behavior. Since the dataset is based on a social setting, it would be misleading if the models yielded different predictions based on different names. To construct the template $\mathbf{T}$, we used the AllenNLP coreference resolution model \cite{gardner2018allennlp}, which has high performance\footnote{F1 score 80.2 on CoNLL benchmark dataset}. This model is used to detect named entities and resolve their corresponding pronouns, facilitating the construction of templates for our experiments.\\\newline
 \begin{table*}[!h]
\centering
\begin{tabular}{ccccc|ccc}
\cline{3-8} &  & \multicolumn{3}{c|}{\begin{tabular}[c]{@{}c@{}}\textbf{Not-finetuned}\\ \textit{(Epoch 0)}\end{tabular}} & \multicolumn{3}{c}{\begin{tabular}[c]{@{}c@{}}\textbf{Finetuned}\\ \textit{(Epoch10)}\end{tabular}} \\ \cline{3-8} 
                                 & & \textsc{GPT2} & \textsc{BERT} & \textsc{RoBERTA}                      & \textsc{GPT2} & \textsc{BERT} & \textsc{RoBERTA}                    \\ \hline
\multirow{4}{*}{Indirect Effect} & \multicolumn{1}{c|}{\textsc{Most}}   & 0.055                      & 0.107                      & 0.074                        & 0.052                     & 0.047                    & 0.037                      \\
                                 & \multicolumn{1}{c|}{\textsc{Least}}  & 0.043                      & 0.091                      & 0.171                        & 0.053                     & 0.039                    & 0.031                      \\ \cline{2-8} 
                                 & \multicolumn{1}{c|}{\textsc{Female}} & 0.072                      & 0.145                      & 0.185                        & 0.079                     & 0.063                    & 0.051                      \\
                                 & \multicolumn{1}{c|}{\textsc{Male}}   & 0.030                      & 0.059                      & 0.034                        & 0.0260                    & 0.025                    & 0.018                      \\ \hline
\end{tabular}
\caption{Indirect Effect of name lists across models. The results show that relative to Non-finetuned models, the indirect effect of names on predictions is marginally reduced in fine-tuned models. }
\label{tab:indirect}
\end{table*}\textbf{Names List} We use U.S. census names dataset\footnote{http://www.ssa.gov/oact/babynames/names.zip}, following \cite{mehrabi2020man} to intervene the name placeholders. It contains 139 years of U.S. census baby names, their corresponding gender, and respective frequencies. To form intervention name lists based on frequency, we filtered out the most frequent $k$ names over all years for $N_{\textsc{Most}}$, and the least frequent $k$ names over all years for $N_{\textsc{Least}}$. We set $k=200$.\\\newline
\textbf{Model} We use three widely used models, \textsc{GPT2} \cite{radford2019language}, \textsc{Bert} \cite{devlin2019bert}, and \textsc{RoBERTa} \cite{liu2019roberta}. We customized each model with a linear layer on top to perform a multiple-choice selection task. The feed-forward (FF) linear layer was obtained by $\text{logits}=\textbf{Model}(X)$, $\hat{y}=\textbf{FF}(\text{logits})$. The hyper-parameter setting for the training is described in Appendix \ref{sec:appendix-a}.

\section{Results and Discussion}
\subsection{Direct Effect}
\textbf{\textsc{Accuracy}} The results of the direct effect of accuracy for different sets of interventions are presented in Table \ref{tab:direct-acc}.  Comparing the first three columns (\textit{not-finetuned}) with the subsequent three columns (\textit{fine-tuned)}, we observe that the causal effect of accuracy is not statistically significant when the models are fine-tuned. This trend holds consistently true across all three models examined in this study. This suggests that the direct effect of name characteristics on accuracy is not significant when fine-tuned. 
The effect sizes of the \textit{not-finetuned} models are reported in accordance with previous literature that predominantly focuses on these models \cite{wolfe2021low,shwartz2020you}. However, it is crucial to emphasize the efficacy of fine-tuning, as it reflects a more realistic scenario for model deployment \cite{jeoung2022changed}. We compared the effect sizes of the not-finetuned models with those of the fine-tuned models, thereby examining the impact of fine-tuning on model behavior. We also provide an analysis of the correlation between the model's accuracy and effect sizes in Appendix \ref{sec:corrl}.
\\\newline
\textbf{\textsc{Agreement}}  The analysis of the direct causal effect of agreement ($\mathbf{d}_{\textsc{AGR}}$) shows that a significant difference in name lists based on frequency persists even after fine-tuning all three models ( Table \ref{tab:direct-agr}, first row). This suggests that despite the fine-tuning process, the models continue to exhibit variations in their agreement on predictions based on the frequency of names used. Specifically, the positive and significant value of $\textsc{Most}\rightarrow \textsc{Least}$ indicates that the prediction is more divergent for\textsc{Least}  than {\textsc{Most}}. This implies that when the model makes incorrect predictions, the resulting predictions tend to be more inconsistent or diverse, rather than consistent.\\
\begin{figure}[]
\centerline{\includegraphics[width=3.5in, height=2.5in]{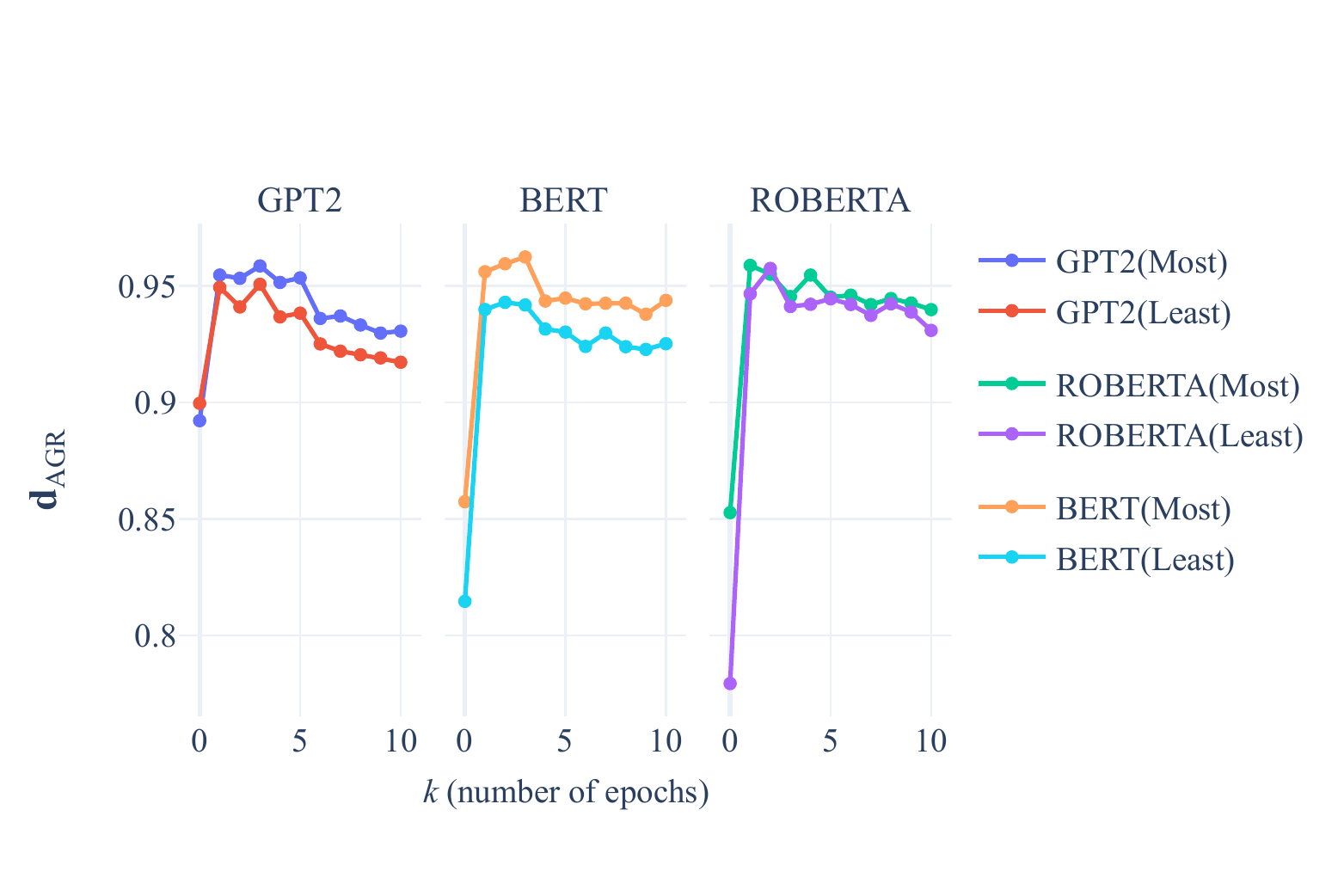}}
\caption{The $\mathbf{d}_\textsc{AGR}$ of \textsc{Most} and \textsc{Least} values over the training phase (number of epochs). For \textsc{GPT2} and \textsc{BERT}, the gap of \textsc{Most} values and \textsc{Least} is consistent across the number of epochs.}
\label{fig:direct_epochs}
\end{figure}
Figure \ref{fig:direct_epochs} illustrates the disentangled values for $\mathbf{d}_\textsc{AGR}$ across different epochs during the training phase. For both \textsc{GPT2} and \textsc{BERT}, a consistent gap between \textsc{Most} and \textsc{Least} is observed throughout the training epochs. In contrast, for \textsc{RoBERTa}, although the gap is not consistent across all epochs, the agreement measures for \textsc{Most} remain consistently higher than those for \textsc{Least}. This discrepancy in the gap between \textsc{RoBERTa} and the other models could potentially be attributed to the robust optimization design of \textsc{RoBERTa}, which complements that of \textsc{BERT} \cite{liu2019roberta}. Also, these findings are consistent with the conclusion drawn by \cite{basu2021but}, who also observed that \textsc{RoBERTa} generates the most robust results. Overall, the findings indicate that the agreement ratio of $\textsc{Least}$ consistently remains lower than that of $\textsc{Most}$ throughout the training phase, suggesting that the predictions for \textsc{Least} are more divergent.
\subsection{Indirect Effect} 
Table \ref{tab:indirect} presents the results pertaining to the indirect effect of name lists on predictions. Specifically, the indirect effect quantifies the sensitivity of pronouns associated with names on model predictions. Overall, the findings indicate that, in comparison to non-finetuned models, the indirect effect of names on predictions is marginally reduced in fine-tuned models. For \textsc{Bert} and \textsc{RoBERTa}, the indirect effect of both frequency and gender is diminished when finetuned. However, for \textsc{Gpt2}, the indirect effect is reduced in most cases, except for the name lists of \textsc{Least} and \textsc{Females}.

\begin{table*}[!ht]
\centering
\begin{tabular}{ll}
\hline & \multicolumn{1}{c}{Examples} \\ \hline
\multicolumn{1}{l|}{\multirow{2}{*}{\rotatebox[origin=c]{90}{\textbf{\textit{Frequent Names}}}}}  & \begin{tabular}[c]{@{}l@{}} {\setlength{\fboxsep}{0pt}\colorbox{white!0}{\parbox{0.9\textwidth}{\ \linebreak \small
\colorbox[RGB]{102,197,204}{\strut Mary} \colorbox[RGB]{135,197,95}{\strut was} \colorbox[RGB]{135,197,95}{\strut always} \colorbox[RGB]{135,197,95}{\strut the} \colorbox[RGB]{135,197,95}{\strut type} \colorbox[RGB]{248,156,116}{\strut who} \colorbox[RGB]{135,197,95}{\strut liked} \colorbox[RGB]{248,156,116}{\strut to} \colorbox[RGB]{246,207,113}{\strut party} \colorbox[RGB]{135,197,95}{\strut ,} \colorbox[RGB]{248,156,116}{\strut she} \colorbox[RGB]{248,156,116}{\strut was} \colorbox[RGB]{246,207,113}{\strut excited} \colorbox[RGB]{246,207,113}{\strut it} \colorbox[RGB]{246,207,113}{\strut was} \colorbox[RGB]{246,207,113}{\strut her} \colorbox[RGB]{246,207,113}{\strut birthday} \colorbox[RGB]{201,219,116}{\strut ,} \colorbox[RGB]{246,207,113}{\strut and} \colorbox[RGB]{246,207,113}{\strut invited} \colorbox[RGB]{246,207,113}{\strut people} \colorbox[RGB]{246,207,113}{\strut to} \colorbox[RGB]{246,207,113}{\strut her} \colorbox[RGB]{246,207,113}{\strut house} \colorbox[RGB]{201,219,116}{\strut .} \colorbox[RGB]{201,219,116}{\strut [SEP]} \colorbox[RGB]{201,219,116}{\strut [SEP]} \colorbox[RGB]{158,185,243}{\strut why} \colorbox[RGB]{158,185,243}{\strut did} \colorbox[RGB]{158,185,243}{\strut Mary} \colorbox[RGB]{158,185,243}{\strut do} \colorbox[RGB]{158,185,243}{\strut this} \colorbox[RGB]{201,219,116}{\strut ?} \colorbox[RGB]{246,207,113}{\strut loved} \colorbox[RGB]{246,207,113}{\strut the} \colorbox[RGB]{246,207,113}{\strut party} \colorbox[RGB]{246,207,113}{\strut scene} \colorbox[RGB]{254,136,177}{\strut [PAD]} \colorbox[RGB]{254,136,177}{\strut [PAD]}  \colorbox[RGB]{248,156,116}{\strut social} \colorbox[RGB]{254,136,177}{\strut ize} \colorbox[RGB]{254,136,177}{\strut [PAD]} \colorbox[RGB]{254,136,177}{\strut [PAD]} \colorbox[RGB]{158,185,243}{\strut was} \colorbox[RGB]{139,224,164}{\strut not} \colorbox[RGB]{139,224,164}{\strut the} \colorbox[RGB]{139,224,164}{\strut party} \colorbox[RGB]{139,224,164}{\strut girl} \colorbox[RGB]{139,224,164}{\strut [PAD]} \colorbox[RGB]{139,224,164}{\strut [PAD]} \colorbox[RGB]{139,224,164}{\strut [PAD]} \ \linebreak 
}}}\\ {\setlength{\fboxsep}{0pt}\colorbox{white!0}{\parbox{0.9\textwidth}{ \small
\colorbox[RGB]{102,197,204}{\strut Elizabeth} \colorbox[RGB]{135,197,95}{\strut was} \colorbox[RGB]{135,197,95}{\strut always} \colorbox[RGB]{135,197,95}{\strut the} \colorbox[RGB]{135,197,95}{\strut type} \colorbox[RGB]{248,156,116}{\strut who} \colorbox[RGB]{135,197,95}{\strut liked} \colorbox[RGB]{248,156,116}{\strut to} \colorbox[RGB]{246,207,113}{\strut party} \colorbox[RGB]{135,197,95}{\strut ,} \colorbox[RGB]{248,156,116}{\strut she} \colorbox[RGB]{248,156,116}{\strut was} \colorbox[RGB]{246,207,113}{\strut excited} \colorbox[RGB]{246,207,113}{\strut it} \colorbox[RGB]{246,207,113}{\strut was} \colorbox[RGB]{246,207,113}{\strut her} \colorbox[RGB]{246,207,113}{\strut birthday} \colorbox[RGB]{201,219,116}{\strut ,} \colorbox[RGB]{246,207,113}{\strut and} \colorbox[RGB]{246,207,113}{\strut invited} \colorbox[RGB]{246,207,113}{\strut people} \colorbox[RGB]{246,207,113}{\strut to} \colorbox[RGB]{246,207,113}{\strut her} \colorbox[RGB]{246,207,113}{\strut house} \colorbox[RGB]{201,219,116}{\strut .} \colorbox[RGB]{201,219,116}{\strut [SEP]} \colorbox[RGB]{201,219,116}{\strut [SEP]} \colorbox[RGB]{158,185,243}{\strut why} \colorbox[RGB]{158,185,243}{\strut did} \colorbox[RGB]{158,185,243}{\strut Elizabeth} \colorbox[RGB]{158,185,243}{\strut do} \colorbox[RGB]{158,185,243}{\strut this} \colorbox[RGB]{201,219,116}{\strut ?} \colorbox[RGB]{246,207,113}{\strut loved} \colorbox[RGB]{246,207,113}{\strut the} \colorbox[RGB]{246,207,113}{\strut party} \colorbox[RGB]{246,207,113}{\strut scene} \colorbox[RGB]{254,136,177}{\strut [PAD]} \colorbox[RGB]{254,136,177}{\strut [PAD]} \colorbox[RGB]{248,156,116}{\strut social} \colorbox[RGB]{254,136,177}{\strut ize} \colorbox[RGB]{254,136,177}{\strut [PAD]} \colorbox[RGB]{254,136,177}{\strut [PAD]} \colorbox[RGB]{158,185,243}{\strut was} \colorbox[RGB]{139,224,164}{\strut not} \colorbox[RGB]{139,224,164}{\strut the} \colorbox[RGB]{139,224,164}{\strut party} \colorbox[RGB]{139,224,164}{\strut girl} \colorbox[RGB]{139,224,164}{\strut [PAD]} \colorbox[RGB]{139,224,164}{\strut [PAD]} \colorbox[RGB]{139,224,164}{\strut [PAD]}  \ \linebreak }}}\end{tabular}    \\
\multicolumn{1}{l|}{}                             & \begin{tabular}[c]{@{}l@{}} {\setlength{\fboxsep}{0pt}\colorbox{white!0}{\parbox{0.9\textwidth}{ \small
\colorbox[RGB]{102,197,204}{\strut James} \colorbox[RGB]{135,197,95}{\strut was} \colorbox[RGB]{135,197,95}{\strut always} \colorbox[RGB]{135,197,95}{\strut the} \colorbox[RGB]{135,197,95}{\strut type} \colorbox[RGB]{248,156,116}{\strut who} \colorbox[RGB]{135,197,95}{\strut liked} \colorbox[RGB]{248,156,116}{\strut to} \colorbox[RGB]{246,207,113}{\strut party} \colorbox[RGB]{135,197,95}{\strut ,} \colorbox[RGB]{248,156,116}{\strut he} \colorbox[RGB]{248,156,116}{\strut was} \colorbox[RGB]{246,207,113}{\strut excited} \colorbox[RGB]{246,207,113}{\strut it} \colorbox[RGB]{246,207,113}{\strut was} \colorbox[RGB]{246,207,113}{\strut his} \colorbox[RGB]{246,207,113}{\strut birthday} \colorbox[RGB]{246,207,113}{\strut ,} \colorbox[RGB]{246,207,113}{\strut and} \colorbox[RGB]{246,207,113}{\strut invited} \colorbox[RGB]{246,207,113}{\strut people} \colorbox[RGB]{246,207,113}{\strut to} \colorbox[RGB]{246,207,113}{\strut his} \colorbox[RGB]{246,207,113}{\strut house} \colorbox[RGB]{201,219,116}{\strut .} \colorbox[RGB]{201,219,116}{\strut [SEP]} \colorbox[RGB]{201,219,116}{\strut [SEP]} \colorbox[RGB]{158,185,243}{\strut why} \colorbox[RGB]{158,185,243}{\strut did} \colorbox[RGB]{158,185,243}{\strut James} \colorbox[RGB]{158,185,243}{\strut do} \colorbox[RGB]{158,185,243}{\strut this} \colorbox[RGB]{158,185,243}{\strut ?} \colorbox[RGB]{246,207,113}{\strut loved} \colorbox[RGB]{246,207,113}{\strut the} \colorbox[RGB]{246,207,113}{\strut party} \colorbox[RGB]{246,207,113}{\strut scene} \colorbox[RGB]{254,136,177}{\strut [PAD]} \colorbox[RGB]{254,136,177}{\strut [PAD]}  \colorbox[RGB]{248,156,116}{\strut social} \colorbox[RGB]{254,136,177}{\strut ize} \colorbox[RGB]{254,136,177}{\strut [PAD]} \colorbox[RGB]{254,136,177}{\strut [PAD]} \colorbox[RGB]{158,185,243}{\strut was} \colorbox[RGB]{139,224,164}{\strut not} \colorbox[RGB]{139,224,164}{\strut the} \colorbox[RGB]{139,224,164}{\strut party} \colorbox[RGB]{139,224,164}{\strut boy} \colorbox[RGB]{139,224,164}{\strut [PAD]} \colorbox[RGB]{139,224,164}{\strut [PAD]} \colorbox[RGB]{139,224,164}{\strut [PAD]} \ \linebreak 
}}}\\ {\setlength{\fboxsep}{0pt}\colorbox{white!0}{\parbox{0.9\textwidth}{ \small
\colorbox[RGB]{102,197,204}{\strut Robert} \colorbox[RGB]{135,197,95}{\strut was} \colorbox[RGB]{135,197,95}{\strut always} \colorbox[RGB]{248,156,116}{\strut the} \colorbox[RGB]{135,197,95}{\strut type} \colorbox[RGB]{248,156,116}{\strut who} \colorbox[RGB]{135,197,95}{\strut liked} \colorbox[RGB]{248,156,116}{\strut to} \colorbox[RGB]{246,207,113}{\strut party} \colorbox[RGB]{135,197,95}{\strut ,} \colorbox[RGB]{248,156,116}{\strut he} \colorbox[RGB]{248,156,116}{\strut was} \colorbox[RGB]{246,207,113}{\strut excited} \colorbox[RGB]{246,207,113}{\strut it} \colorbox[RGB]{246,207,113}{\strut was} \colorbox[RGB]{246,207,113}{\strut his} \colorbox[RGB]{246,207,113}{\strut birthday} \colorbox[RGB]{246,207,113}{\strut ,} \colorbox[RGB]{246,207,113}{\strut and} \colorbox[RGB]{246,207,113}{\strut invited} \colorbox[RGB]{246,207,113}{\strut people} \colorbox[RGB]{246,207,113}{\strut to} \colorbox[RGB]{246,207,113}{\strut his} \colorbox[RGB]{246,207,113}{\strut house} \colorbox[RGB]{201,219,116}{\strut .} \colorbox[RGB]{201,219,116}{\strut [SEP]} \colorbox[RGB]{201,219,116}{\strut [SEP]} \colorbox[RGB]{158,185,243}{\strut why} \colorbox[RGB]{158,185,243}{\strut did} \colorbox[RGB]{158,185,243}{\strut Robert} \colorbox[RGB]{158,185,243}{\strut do} \colorbox[RGB]{158,185,243}{\strut this} \colorbox[RGB]{201,219,116}{\strut ?} \colorbox[RGB]{246,207,113}{\strut loved} \colorbox[RGB]{246,207,113}{\strut the} \colorbox[RGB]{246,207,113}{\strut party} \colorbox[RGB]{246,207,113}{\strut scene} \colorbox[RGB]{254,136,177}{\strut [PAD]} \colorbox[RGB]{254,136,177}{\strut [PAD]}  \colorbox[RGB]{248,156,116}{\strut social} \colorbox[RGB]{254,136,177}{\strut ize} \colorbox[RGB]{254,136,177}{\strut [PAD]} \colorbox[RGB]{254,136,177}{\strut [PAD]} \colorbox[RGB]{158,185,243}{\strut was} \colorbox[RGB]{139,224,164}{\strut not} \colorbox[RGB]{139,224,164}{\strut the} \colorbox[RGB]{139,224,164}{\strut party} \colorbox[RGB]{139,224,164}{\strut boy} \colorbox[RGB]{139,224,164}{\strut [PAD]} \colorbox[RGB]{139,224,164}{\strut [PAD]} \colorbox[RGB]{139,224,164}{\strut [PAD]} \ \linebreak 
}}}\end{tabular}    \\ \hline
\multicolumn{1}{l|}{\multirow{2}{*}{\rotatebox[origin=c]{90}{\textbf{\textit{Less Frequent Names}}}}} & \begin{tabular}[c]{@{}l@{}}{ \setlength{\fboxsep}{0pt}\colorbox{white!0}{\parbox{0.9\textwidth}{\ \linebreak \small
\colorbox[RGB]{102,197,204}{\strut And} \colorbox[RGB]{135,197,95}{\strut rine} \colorbox[RGB]{135,197,95}{\strut was} \colorbox[RGB]{135,197,95}{\strut always} \colorbox[RGB]{135,197,95}{\strut the} \colorbox[RGB]{135,197,95}{\strut type} \colorbox[RGB]{248,156,116}{\strut who} \colorbox[RGB]{135,197,95}{\strut liked} \colorbox[RGB]{248,156,116}{\strut to} \colorbox[RGB]{246,207,113}{\strut party} \colorbox[RGB]{135,197,95}{\strut ,} \colorbox[RGB]{248,156,116}{\strut she} \colorbox[RGB]{248,156,116}{\strut was} \colorbox[RGB]{246,207,113}{\strut excited} \colorbox[RGB]{246,207,113}{\strut it} \colorbox[RGB]{246,207,113}{\strut was} \colorbox[RGB]{246,207,113}{\strut her} \colorbox[RGB]{246,207,113}{\strut birthday} \colorbox[RGB]{201,219,116}{\strut ,} \colorbox[RGB]{246,207,113}{\strut and} \colorbox[RGB]{246,207,113}{\strut invited} \colorbox[RGB]{246,207,113}{\strut people} \colorbox[RGB]{246,207,113}{\strut to} \colorbox[RGB]{246,207,113}{\strut her} \colorbox[RGB]{246,207,113}{\strut house} \colorbox[RGB]{201,219,116}{\strut .} \colorbox[RGB]{201,219,116}{\strut [SEP]} \colorbox[RGB]{201,219,116}{\strut [SEP]} \colorbox[RGB]{158,185,243}{\strut why} \colorbox[RGB]{158,185,243}{\strut did} \colorbox[RGB]{135,197,95}{\strut And} \colorbox[RGB]{158,185,243}{\strut rine} \colorbox[RGB]{158,185,243}{\strut do} \colorbox[RGB]{158,185,243}{\strut this} \colorbox[RGB]{158,185,243}{\strut ?} \colorbox[RGB]{246,207,113}{\strut loved} \colorbox[RGB]{246,207,113}{\strut the} \colorbox[RGB]{246,207,113}{\strut party} \colorbox[RGB]{246,207,113}{\strut scene} \colorbox[RGB]{254,136,177}{\strut [PAD]} \colorbox[RGB]{254,136,177}{\strut [PAD]} \colorbox[RGB]{248,156,116}{\strut social} \colorbox[RGB]{254,136,177}{\strut ize} \colorbox[RGB]{254,136,177}{\strut [PAD]} \colorbox[RGB]{254,136,177}{\strut [PAD]} \colorbox[RGB]{158,185,243}{\strut was} \colorbox[RGB]{139,224,164}{\strut not} \colorbox[RGB]{139,224,164}{\strut the} \colorbox[RGB]{139,224,164}{\strut party} \colorbox[RGB]{139,224,164}{\strut girl} \colorbox[RGB]{139,224,164}{\strut [PAD]} \colorbox[RGB]{139,224,164}{\strut [PAD]} \colorbox[RGB]{139,224,164}{\strut [PAD]} \ \linebreak 
}}}\\ {\setlength{\fboxsep}{0pt}\colorbox{white!0}{\parbox{0.9\textwidth}{  \small
\colorbox[RGB]{102,197,204}{\strut Le} \colorbox[RGB]{135,197,95}{\strut u} \colorbox[RGB]{135,197,95}{\strut ven} \colorbox[RGB]{135,197,95}{\strut ia} \colorbox[RGB]{135,197,95}{\strut was} \colorbox[RGB]{135,197,95}{\strut always} \colorbox[RGB]{135,197,95}{\strut the} \colorbox[RGB]{135,197,95}{\strut type} \colorbox[RGB]{248,156,116}{\strut who} \colorbox[RGB]{135,197,95}{\strut liked} \colorbox[RGB]{248,156,116}{\strut to} \colorbox[RGB]{246,207,113}{\strut party} \colorbox[RGB]{135,197,95}{\strut ,} \colorbox[RGB]{248,156,116}{\strut she} \colorbox[RGB]{248,156,116}{\strut was} \colorbox[RGB]{246,207,113}{\strut excited} \colorbox[RGB]{246,207,113}{\strut it} \colorbox[RGB]{246,207,113}{\strut was} \colorbox[RGB]{246,207,113}{\strut her} \colorbox[RGB]{246,207,113}{\strut birthday} \colorbox[RGB]{246,207,113}{\strut ,} \colorbox[RGB]{246,207,113}{\strut and} \colorbox[RGB]{246,207,113}{\strut invited} \colorbox[RGB]{246,207,113}{\strut people} \colorbox[RGB]{246,207,113}{\strut to} \colorbox[RGB]{246,207,113}{\strut her} \colorbox[RGB]{246,207,113}{\strut house} \colorbox[RGB]{201,219,116}{\strut .} \colorbox[RGB]{201,219,116}{\strut [SEP]} \colorbox[RGB]{201,219,116}{\strut [SEP]} \colorbox[RGB]{158,185,243}{\strut why} \colorbox[RGB]{158,185,243}{\strut did} \colorbox[RGB]{158,185,243}{\strut Le} \colorbox[RGB]{248,156,116}{\strut u} \colorbox[RGB]{248,156,116}{\strut ven} \colorbox[RGB]{158,185,243}{\strut ia} \colorbox[RGB]{158,185,243}{\strut do} \colorbox[RGB]{158,185,243}{\strut this} \colorbox[RGB]{201,219,116}{\strut ?} \colorbox[RGB]{246,207,113}{\strut loved} \colorbox[RGB]{246,207,113}{\strut the} \colorbox[RGB]{246,207,113}{\strut party} \colorbox[RGB]{246,207,113}{\strut scene} \colorbox[RGB]{254,136,177}{\strut [PAD]} \colorbox[RGB]{248,156,116}{\strut social} \colorbox[RGB]{254,136,177}{\strut ize} \colorbox[RGB]{254,136,177}{\strut [PAD]} \colorbox[RGB]{254,136,177}{\strut [PAD]} \colorbox[RGB]{158,185,243}{\strut was} \colorbox[RGB]{139,224,164}{\strut not} \colorbox[RGB]{139,224,164}{\strut the} \colorbox[RGB]{139,224,164}{\strut party} \colorbox[RGB]{139,224,164}{\strut girl} \ \linebreak 
}}}\end{tabular}  \\
\multicolumn{1}{l|}{}                             & \begin{tabular}[c]{@{}l@{}} {\setlength{\fboxsep}{0pt}\colorbox{white!0}{\parbox{0.9\textwidth}{ \small
\colorbox[RGB]{102,197,204}{\strut Nav} \colorbox[RGB]{135,197,95}{\strut ajo} \colorbox[RGB]{135,197,95}{\strut was} \colorbox[RGB]{135,197,95}{\strut always} \colorbox[RGB]{135,197,95}{\strut the} \colorbox[RGB]{135,197,95}{\strut type} \colorbox[RGB]{248,156,116}{\strut who} \colorbox[RGB]{135,197,95}{\strut liked} \colorbox[RGB]{248,156,116}{\strut to} \colorbox[RGB]{246,207,113}{\strut party} \colorbox[RGB]{135,197,95}{\strut ,} \colorbox[RGB]{246,207,113}{\strut he} \colorbox[RGB]{248,156,116}{\strut was} \colorbox[RGB]{246,207,113}{\strut excited} \colorbox[RGB]{246,207,113}{\strut it} \colorbox[RGB]{246,207,113}{\strut was} \colorbox[RGB]{246,207,113}{\strut his} \colorbox[RGB]{246,207,113}{\strut birthday} \colorbox[RGB]{246,207,113}{\strut ,} \colorbox[RGB]{246,207,113}{\strut and} \colorbox[RGB]{246,207,113}{\strut invited} \colorbox[RGB]{246,207,113}{\strut people} \colorbox[RGB]{246,207,113}{\strut to} \colorbox[RGB]{246,207,113}{\strut his} \colorbox[RGB]{246,207,113}{\strut house} \colorbox[RGB]{201,219,116}{\strut .} \colorbox[RGB]{201,219,116}{\strut [SEP]} \colorbox[RGB]{201,219,116}{\strut [SEP]} \colorbox[RGB]{158,185,243}{\strut why} \colorbox[RGB]{158,185,243}{\strut did} \colorbox[RGB]{246,207,113}{\strut Navajo} \colorbox[RGB]{158,185,243}{\strut do} \colorbox[RGB]{158,185,243}{\strut this} \colorbox[RGB]{201,219,116}{\strut ?} \colorbox[RGB]{246,207,113}{\strut loved} \colorbox[RGB]{246,207,113}{\strut the} \colorbox[RGB]{246,207,113}{\strut party} \colorbox[RGB]{246,207,113}{\strut scene} \colorbox[RGB]{254,136,177}{\strut [PAD]} \colorbox[RGB]{254,136,177}{\strut [PAD]} \colorbox[RGB]{248,156,116}{\strut social} \colorbox[RGB]{248,156,116}{\strut ize} \colorbox[RGB]{254,136,177}{\strut [PAD]} \colorbox[RGB]{254,136,177}{\strut [PAD]} \colorbox[RGB]{254,136,177}{\strut [PAD]} \colorbox[RGB]{158,185,243}{\strut was} \colorbox[RGB]{139,224,164}{\strut not} \colorbox[RGB]{139,224,164}{\strut the} \colorbox[RGB]{139,224,164}{\strut party} \colorbox[RGB]{139,224,164}{\strut girl} \colorbox[RGB]{139,224,164}{\strut [PAD]} \colorbox[RGB]{139,224,164}{\strut [PAD]} \colorbox[RGB]{139,224,164}{\strut [PAD]} \ \linebreak 
}}}\\ {\setlength{\fboxsep}{0pt}\colorbox{white!0}{\parbox{0.9\textwidth}{ \small
\colorbox[RGB]{102,197,204}{\strut Wind} \colorbox[RGB]{135,197,95}{\strut field} \colorbox[RGB]{135,197,95}{\strut was} \colorbox[RGB]{135,197,95}{\strut always} \colorbox[RGB]{135,197,95}{\strut the} \colorbox[RGB]{135,197,95}{\strut type} \colorbox[RGB]{248,156,116}{\strut who} \colorbox[RGB]{135,197,95}{\strut liked} \colorbox[RGB]{248,156,116}{\strut to} \colorbox[RGB]{246,207,113}{\strut party} \colorbox[RGB]{135,197,95}{\strut ,} \colorbox[RGB]{248,156,116}{\strut he} \colorbox[RGB]{248,156,116}{\strut was} \colorbox[RGB]{246,207,113}{\strut excited} \colorbox[RGB]{246,207,113}{\strut it} \colorbox[RGB]{246,207,113}{\strut was} \colorbox[RGB]{246,207,113}{\strut his} \colorbox[RGB]{246,207,113}{\strut birthday} \colorbox[RGB]{246,207,113}{\strut ,} \colorbox[RGB]{246,207,113}{\strut and} \colorbox[RGB]{246,207,113}{\strut invited} \colorbox[RGB]{246,207,113}{\strut people} \colorbox[RGB]{246,207,113}{\strut to} \colorbox[RGB]{246,207,113}{\strut his} \colorbox[RGB]{246,207,113}{\strut house} \colorbox[RGB]{201,219,116}{\strut .} \colorbox[RGB]{201,219,116}{\strut [SEP]} \colorbox[RGB]{201,219,116}{\strut [SEP]} \colorbox[RGB]{158,185,243}{\strut why} \colorbox[RGB]{158,185,243}{\strut did} \colorbox[RGB]{158,185,243}{\strut Wind} \colorbox[RGB]{158,185,243}{\strut field} \colorbox[RGB]{158,185,243}{\strut do} \colorbox[RGB]{158,185,243}{\strut this} \colorbox[RGB]{158,185,243}{\strut ?}  \colorbox[RGB]{246,207,113}{\strut loved} \colorbox[RGB]{246,207,113}{\strut the} \colorbox[RGB]{246,207,113}{\strut party} \colorbox[RGB]{246,207,113}{\strut scene} \colorbox[RGB]{254,136,177}{\strut [PAD]} \colorbox[RGB]{254,136,177}{\strut [PAD]} \colorbox[RGB]{248,156,116}{\strut social} \colorbox[RGB]{254,136,177}{\strut ize} \colorbox[RGB]{254,136,177}{\strut [PAD]} \colorbox[RGB]{254,136,177}{\strut [PAD]} \colorbox[RGB]{158,185,243}{\strut was} \colorbox[RGB]{139,224,164}{\strut not} \colorbox[RGB]{139,224,164}{\strut the} \colorbox[RGB]{139,224,164}{\strut party} \colorbox[RGB]{139,224,164}{\strut girl} \colorbox[RGB]{139,224,164}{\strut [PAD]} \colorbox[RGB]{139,224,164}{\strut [PAD]} \colorbox[RGB]{139,224,164}{\strut [PAD]} \ \linebreak 
}}} \end{tabular} \\ \hline
\end{tabular}
\caption{Neuron Activation analysis. The section above lists the examples of Frequent Names: \textit{Mary, Elizabeth, James, Robert} while the section below shows the examples of Least Frequent Names: \textit{Andrine, Leuven, Navajo, Windfield}. The color corresponds to the group of components of the neurons that are activated.} 
\label{tab:example_neuron_activation}
\end{table*}

\subsection{Contextualization Measures}
\begin{figure}[]
\centerline{\includegraphics[width=3in, height=2in]{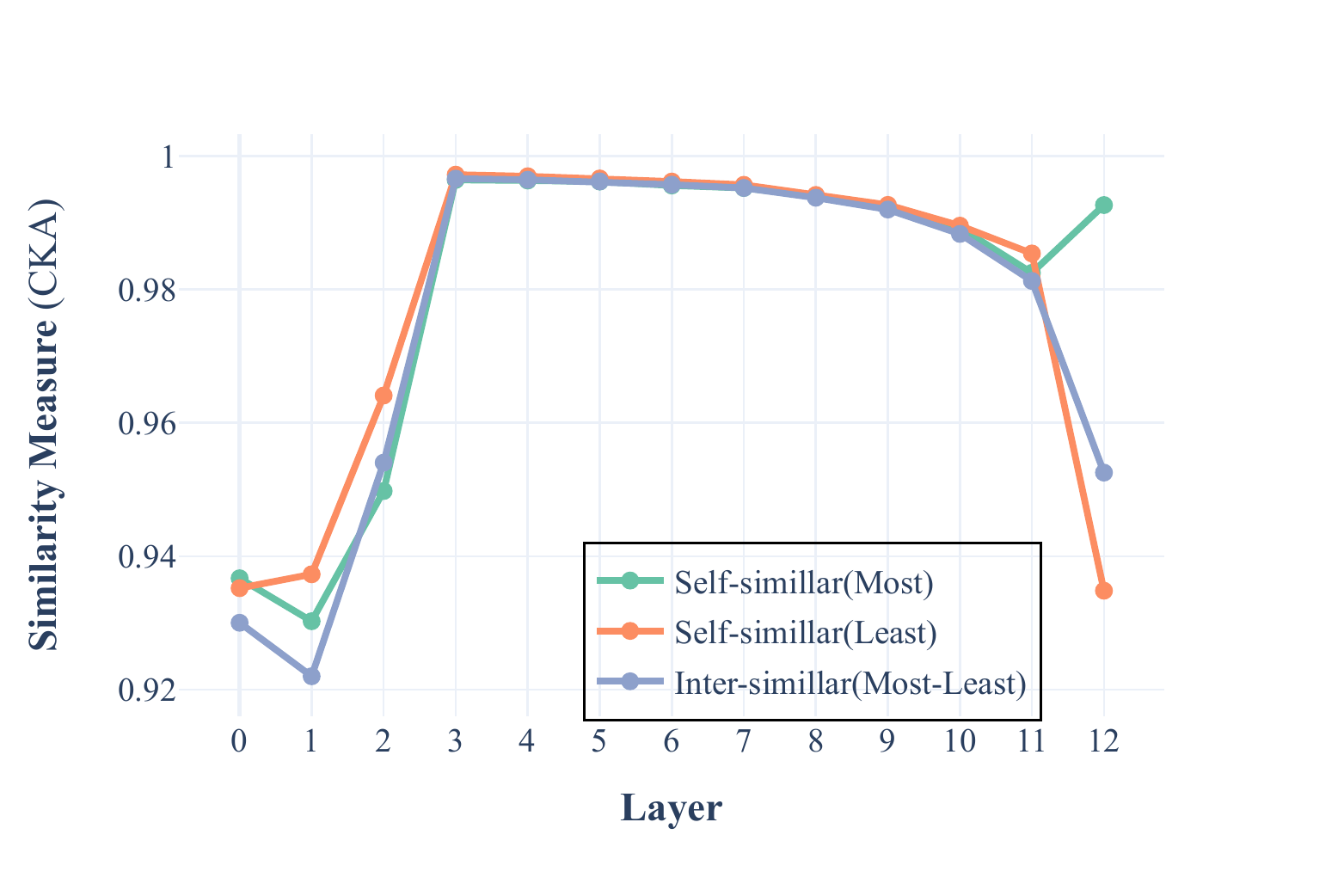}}
\caption{CKA measures across layers}
\label{fig:CKA}
\end{figure}

\begin{figure}[]
\centerline{\includegraphics[width=3in, height=2in]{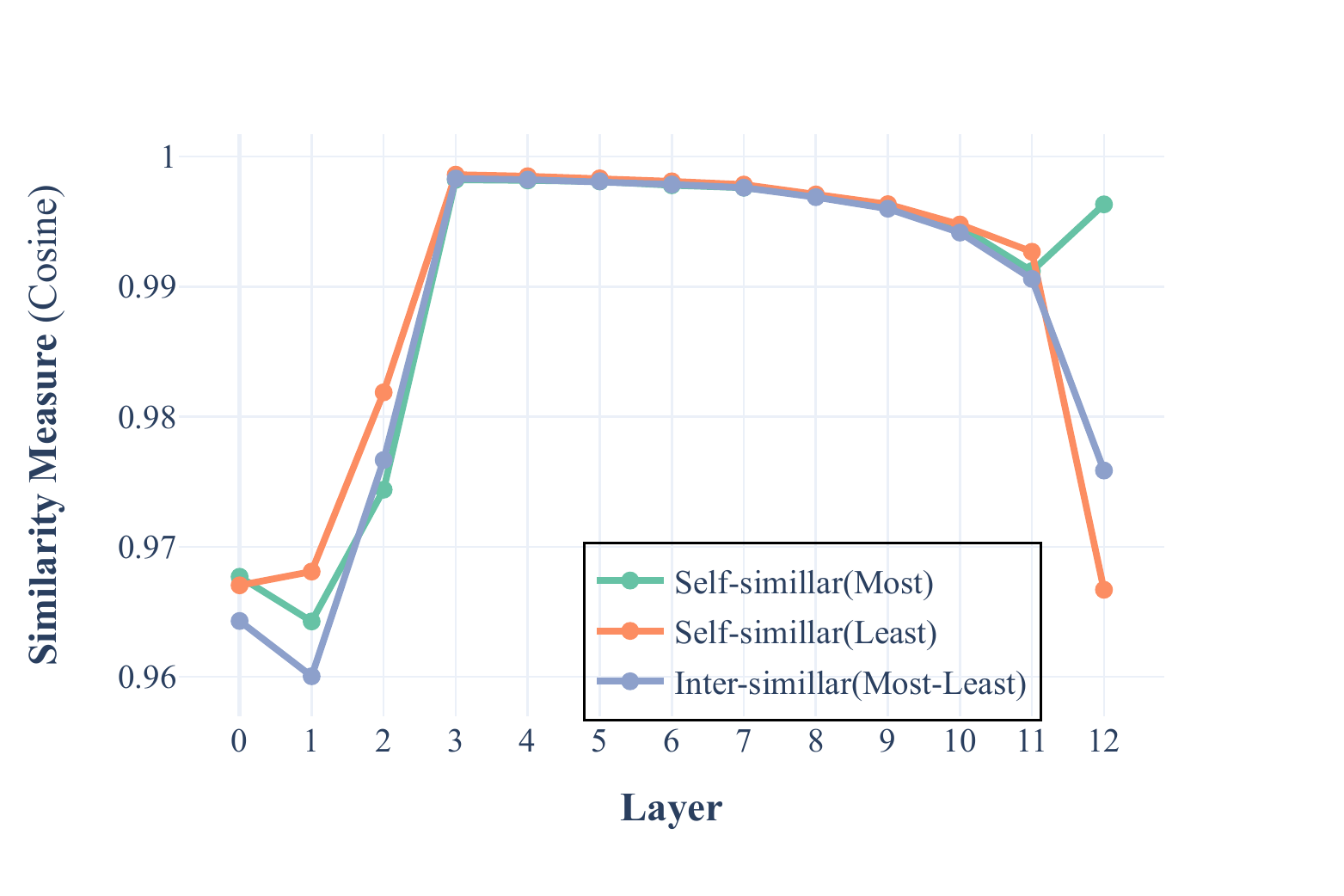}}
\caption{Cosine similarity measures across layers}
\label{fig:Cosine}
\end{figure}

In order to gain insight into how names are internally contextualized in the transformer models, we conducted a preliminary analysis of name representations. To do so, we extracted the embeddings of $N_{\textsc{Most}}$ and $N_{\textsc{Least}}$ samples from fine-tuned GPT2 and measured their similarity. The results are presented in Figure \ref{fig:CKA} and \ref{fig:Cosine}. The \textsc{Self-similar}(Most) and \textsc{Self-similar}(Least) measures represent the similarity between the \textsc{Most} and \textsc{Least} names, respectively, while the \textsc{Inter-similarity}(Most-Least) measure quantifies the similarity between the Most and Least names. The trends observed for both CKA and cosine similarity measures are similar, although with different magnitudes (details of these metrics are discussed in section \ref{sec:Explanation}). These consistent trends are robust across different evaluation metrics. The results show that in the first two layers, the similarity scores are low, but they increase across the mid-layers. However, in the last layer, the similarity of the embeddings of \textsc{Least} names is lower compared to \textsc{Most} names. This finding partly explains Table \ref{tab:direct-agr} first row, which indicates the fine-tuned GPT2 has a significant direct effect on the agreement measure on \textsc{Most} and \textsc{Least}. The relatively low similarity of the embeddings of \textsc{Least} names shows that they exhibit higher variability, being less contextualized compared to that of \textsc{Most}.

\subsection{Neuron Activations}
To further investigate the differences in neuron activations, we conducted an analysis using GPT2 fine-tuned model. The results of this analysis are presented in Table \ref{tab:example_neuron_activation}, where each color represents the components of the neurons that got activated. These components correspond to the clusters obtained from the non-negative factorization on feed-forward neurons. Our observations indicate that less frequent names exhibit two distinct behaviors: 1) they are sub-tokenized into two or more tokens, and 2) they are not activated by the same neuron components as the frequent names. This analysis does not provide an explanation for the \textit{cause} or \textit{reason} for the divergent predictions but rather sheds light on the internal behavior of the model, namely how the neurons activate, which may be related to the divergent predictions observed for the least frequent names. 

\subsection{Mitigating Strategy: Data Augmentation}
Our findings suggest that incorporating a more diverse set of first names into the training data can serve as a potential approach to mitigate the divergent behavior of language models. 
Among all first names in the \textsc{SocialIQA} training dataset, we observed around 66\% of first name instances represent the 10\%  of the most frequent first names in the U.S. Census data. In terms of frequency, these names account for 97\% of all first-name instances in the training dataset (Fig in Appendix \ref{Appendix: train_count}). Such skewed yet highly likely distributions of demographic information in the training dataset may inadvertently introduce biases in the model outputs, as evidenced by previous studies \cite{buolamwini2018gender, karkkainen2021fairface}. To address this issue, recent research by \cite{qian2022perturbation} has demonstrated that augmenting the training data with diverse social demographics can lead to improved model performance and robustness. 

\section{Related Work} 
Previous research has shown that pre-trained language models are susceptible to biases related to people's first names, e.g., in the contexts of sentiment analysis \cite{prabhakaran2019perturbation} and text generation \cite{shwartz2020you}. \citet{wolfe2021low} demonstrated that less common names are more likely to be subtokenized and associated with negative sentiments compared to frequent names. 
In our work, we further extended this prior work by analyzing the impact of fine-tuning models on first names adopting the causal framework.\\
A growing body of research has explored the incorporation of causality in language models. For instance, \citet{feder2021causalm} proposed a causal framework by incorporating additional fine-tuning on adversarial tasks. Similarly, \citet{vig2020investigating} demonstrated the use of causal mediation on language models to mitigate gender bias. Unlike \citet{vig2020investigating}, our approach focuses on applying causal analysis in the input sequence space and exploring the causal relationships between input sequence components and model predictions. 


\section{Conclusion}
In this paper, we introduced a controlled experimental framework to assess the causal effect of first names on commonsense reasoning. Our findings show that the frequency of first names exerts a direct impact on model predictions, with less frequent names leading to divergent outcomes. We suggest careful consideration of the demographics in dataset design.  
\section{Broader Impact}
The data used in our analysis contains no private user information. As for ethical impact, the systematic experimental design we used provides an approach for conducting controlled experiments in the context of natural language processing research, particularly with a focus on the influence of first names on language models.
\section{Limitation}
Our investigation focuses on one aspect of commonsense reasoning restricted to one dataset. There may be numerous other factors in real-world applications. Therefore, our findings may not comprehensively capture the entirety of commonsense reasoning phenomena. Another limitation is that for the sake of simplicity and feasibility, we assumed a fixed threshold of k=200 to categorize frequent and less frequent names. However, this threshold may not be universally applicable to all contexts or datasets, and different thresholds could lead to different results.\\
 

%
\bibliography{acl_latex}
\bibliographystyle{acl_natbib}
%

\appendix

\section{Training Hyperparameters}
\label{sec:appendix-a}
For the train/test split, we followed the original split provided by the data source \cite{sap2019social}. The hyper-parameters used for training are as follows: \textit{AdamW} optimizer, with learning rate \textit{1e-5}, 10 epochs. The checkpoints were saved at the end of every epoch. 
\\
\\
\\\newline
\begin{table*}[ht]
{
\centering
\begin{tabular}{lccc|ccc}
\hline
\multicolumn{1}{c|}{\multirow{2}{*}{\begin{tabular}[c]{@{}c@{}}Effect size:\\ Model\end{tabular}}} 
  & \multicolumn{3}{c|}{\begin{tabular}[c]{@{}c@{}} $\mathbf{d}_{ACC}$\end{tabular}} & \multicolumn{3}{c}{\begin{tabular}[c]{@{}c@{}}$\mathbf{d}_{AGR}$\end{tabular}}\\ \cline{2-7} \multicolumn{1}{l|}{} 
 & \textsc{Gpt2}  & \textsc{Bert}  & \textsc{RoBERTa}                              & \textsc{Gpt2}  & \textsc{Bert} & \textsc{RoBERTa}                                                           \\ \hline
\multicolumn{1}{c|}{\textsc{Most} $\rightarrow$ \textsc{Least}} & \begin{tabular}[c]{@{}c@{}}.473\\ $^{(.142)}$\end{tabular}   & \begin{tabular}[c]{@{}c@{}}-.427\\ $^{(0.19)}$\end{tabular} & \begin{tabular}[c]{@{}c@{}}-.109\\ $^{(.75)}$\end{tabular}                & \begin{tabular}[c]{@{}c@{}}.536\\ $^{(.089)}$\end{tabular}           & \begin{tabular}[c]{@{}c@{}}-.500\\ $^{(.117)}$\end{tabular}              & \begin{tabular}[c]{@{}c@{}}.500\\ $^{(.17)}$\end{tabular} \\
\multicolumn{1}{c|}{\textsc{Male} $\rightarrow$ \textsc{Female}}              & \begin{tabular}[c]{@{}c@{}}.045\\ $^{(.894)}$\end{tabular}   & \begin{tabular}[c]{@{}c@{}}-.264\\ $^{(.433)}$\end{tabular}               & \begin{tabular}[c]{@{}c@{}}.055\\ $^{(.873)}$\end{tabular}               & \begin{tabular}[c]{@{}c@{}}-.555\\ $^{(.077)}$\end{tabular}           & \begin{tabular}[c]{@{}c@{}}-.55\\ $^{(.077)}$\end{tabular}             & \begin{tabular}[c]{@{}c@{}}-.591\\ $^{(.056)}$\end{tabular}           \\
\multicolumn{1}{c|}{\begin{tabular}[c]{@{}l@{}}\textsc{Most Male} $\rightarrow$ \textsc{Least Male}\end{tabular}}     & \begin{tabular}[c]{@{}c@{}}.264\\ $^{(.433)}$\end{tabular}  & \begin{tabular}[c]{@{}c@{}}-$\mathbf{.609}^{*}$\\ $^{(.047)}$\end{tabular}  & \begin{tabular}[c]{@{}c@{}}-.073\\ $^{(.832)}$\end{tabular}               & \begin{tabular}[c]{@{}c@{}}.473\\ $^{(.142)}$\end{tabular}          & \begin{tabular}[c]{@{}c@{}}-$\mathbf{.645}^{*}$\\ $^{(.032)}$\end{tabular}              & \begin{tabular}[c]{@{}c@{}}-.6\\ $^{(.051)}$\end{tabular} \\
\multicolumn{1}{c|}{\begin{tabular}[c]{@{}l@{}}\textsc{Most Female} $\rightarrow$ \textsc{Least Female}\end{tabular}} & \begin{tabular}[c]{@{}c@{}}$\mathbf{.618}^{*}$\\ $^{(.043)}$\end{tabular}  & \begin{tabular}[c]{@{}c@{}}-.264\\ $^{(.433)}$\end{tabular}  & \begin{tabular}[c]{@{}c@{}}.191\\ $^{(.574)}$\end{tabular}                & \begin{tabular}[c]{@{}c@{}}.391\\ $^{(.235)}$\end{tabular}           & \begin{tabular}[c]{@{}c@{}}-.418\\ $^{(.201)}$\end{tabular}              & \begin{tabular}[c]{@{}c@{}}-.145\\ $^{(.67)}$\end{tabular}    \\\hline
\end{tabular}}
\caption{Spearman Correlation between Model's Accuracy and Effect Size: The values show the Spearman's Correlation between the model's accuracy with the effect size ($\mathbf{d_{\textsc{Acc}}}$ and $\mathbf{d_{\textsc{AGR}}}$). The numbers in parentheses indicate the \textit{p-values}. The values in bold indicate the statistical significance with \textit{p-values}$<0.05$. The results show that in most cases, the correlation values are not statistically significant. }
\label{tab:correlation}
\end{table*}

\section{Neuron Activation Analysis}
\label{sec:algorithm}
\SetKwComment{Comment}{/* }{ */}
 \RestyleAlgo{ruled}
 \begin{algorithm}
 \caption{Neuron Activation Analysis}\label{alg:algorithm}
 \KwData{$\mathbf{X}:=(x_1,x_2,\dots,x_n)$, $n$ tokens} 
 \KwResult{$\mathbf{M} \in \mathbb{R}^{k \times n}$, $k$ components}
 $L\gets \text{\# layers}$\;
 \For{$i\gets1$ \KwTo L}{ 
 $\mathbf{X'}\gets f \textbf{block}_i(\text{pre-mlp}_i(\mathbf{X}))$\;
 $\mathbf{y}_i\gets f \textbf{block}_i{\text{mlp}}(\mathbf{X'})$
 }
 $\mathbf{Y}\gets concat(\mathbf{y}_1,\mathbf{y}_2,\dots \mathbf{y}_L)$ $\in \mathbb{R}^{L\times h\times n}$\;
 $\mathbf{M}\gets \text{NMF}(\mathbf{Y})$
 \end{algorithm}
\newpage
 \section{\textsc{SocialIQA} train set names configuration}
 \label{Appendix: train_count}
 \begin{figure}[!h]
\centerline{\includegraphics[width=3in,height=2in]{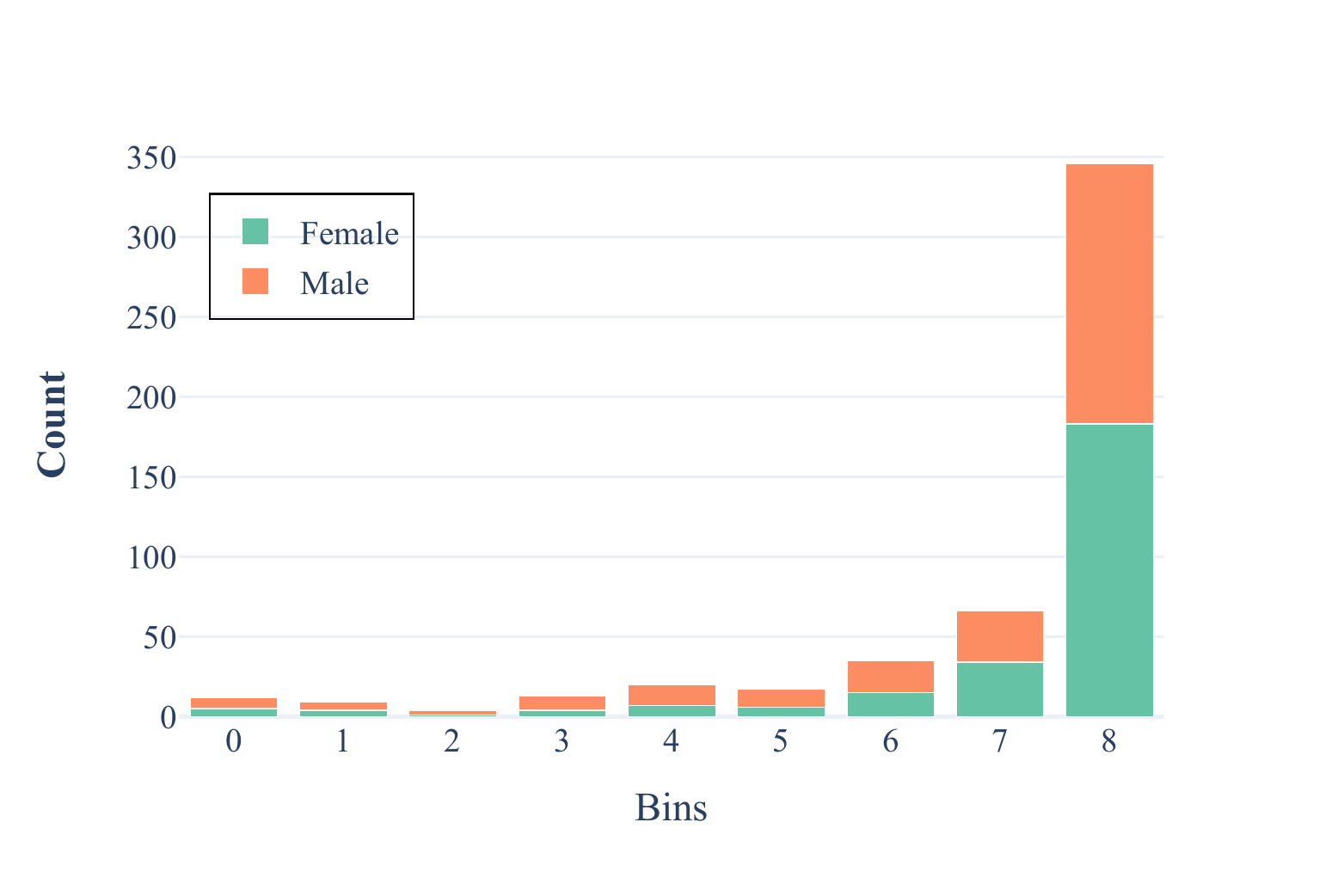}}
\includegraphics[width=3in,height=2in]{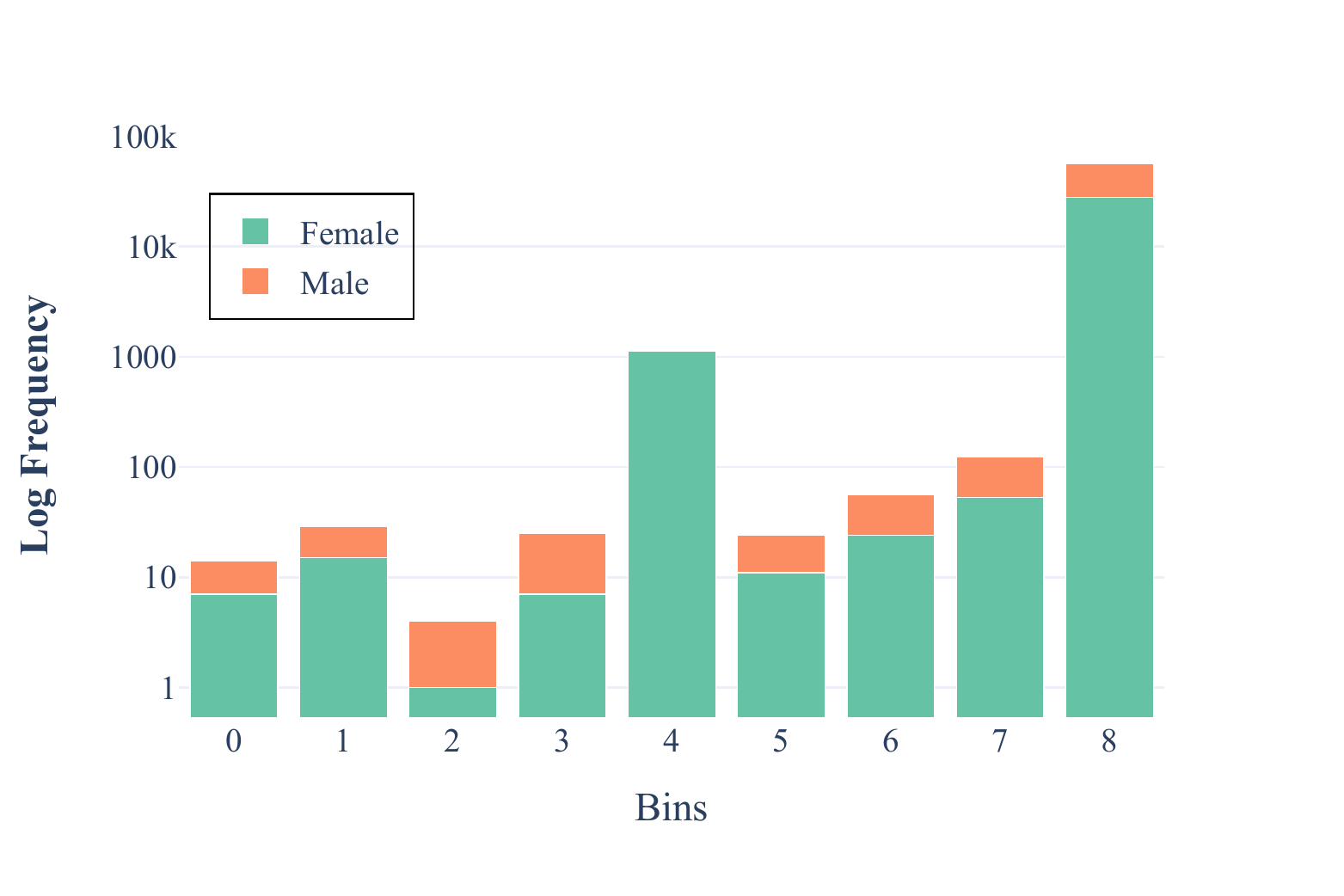}
\caption{Distribution of first names in the train split in \textsc{SocialIQA} dataset. The first names are sorted in ascending order based on U.S. census data frequency and filled into the bins based on quantiles. The $x$-axis represents the Bins. (Above) displays the count of the first names that fall into those bins, showing the prevalence of first names based on whether they are used in the training set of not (Below) shows the frequency of these names in the dataset on a logarithmic scale along the $y$-axis, showing how frequently these names appear in the dataset.}
\label{fig:train_count}
\end{figure}
\newpage
\section{Accuracy and Effect Size Correlation analysis}
\label{sec:corrl}
The relationship between the effect size and the model's performance, measured by accuracy, was investigated in order to determine whether there was any correlation. Table \ref{tab:correlation} presents the correlation analysis between the model's accuracy and two corresponding effect sizes, namely ($\mathbf{d}_{ACC}$, and $\mathbf{d}_{AGR}$). Specifically, for each epoch during the fine-tuning phase, the model's accuracy and effect sizes were compared, and Spearman's correlation coefficient was computed. The results indicate that, in most cases, the correlation values were not statistically significant (\textit{p values} $\leq 0.05$). This suggests that there is no significant association between the improvement in model accuracy and corresponding effect sizes, either positive or negative. By examining the raw data, it was observed that while the models' accuracy increased, the effect sizes remained relatively constant (as shown in Fig \ref{fig:direct_epochs}) throughout some points of the epoch, indicating that there exists some bottleneck in fine-tuning process, as the effect sizes were not effectively mitigated even with the improvement in accuracy.

\end{document}